%% file: main.tex
\documentclass[10pt,twocolumn,letterpaper]{article}

\usepackage{cvpr}
\usepackage{times}
\usepackage{epsfig}
\usepackage{graphicx}
\usepackage{amsmath}
\usepackage{amssymb}
\usepackage{todonotes}
\usepackage{booktabs}
\usepackage{multirow}
\usepackage[font={footnotesize}]{caption}
\usepackage{flushend}
\usepackage[accsupp]{axessibility}

\usepackage{hyperref}
\hypersetup{
    colorlinks=true,
    linkcolor=blue,
    filecolor=magenta,      
    urlcolor=cyan,
    pdftitle={Overleaf Example},
    pdfpagemode=FullScreen,
}

\begin{document}

\title{On the Biometric Capacity of Generative Face Models}

\author{Vishnu Naresh Boddeti \quad\quad Gautam Sreekumar \quad\quad Arun Ross\\
{\normalsize Michigan State University} \\
{\tt\small \{vishnu, sreekum1, rossarun\}@msu.edu}
}

\maketitle
\thispagestyle{empty}

\input{01-abstract}
\input{02-introduction}
\input{03-related-work}
\input{04-approach}
\input{05-experiments}
\input{06-conclusion}

{\small
\bibliographystyle{ieee}
\bibliography{egbib}
}

\end{document}

%% file: 01-abstract.tex
\begin{abstract}
There has been tremendous progress in generating realistic faces with high fidelity over the past few years. Despite this progress, a crucial question remains unanswered: ``Given a generative face model, how many unique identities can it generate?" In other words, what is the biometric capacity of the generative face model? A scientific basis for answering this question will benefit evaluating and comparing different generative face models and establish an upper bound on their scalability. This paper proposes a statistical approach to estimate the biometric capacity of generated face images in a hyperspherical feature space. We employ our approach on multiple generative models, including unconditional generators like StyleGAN, Latent Diffusion Model, and ``Generated Photos," as well as DCFace, a class-conditional generator. We also estimate capacity w.r.t. demographic attributes such as gender and age. Our capacity estimates indicate that (a) under ArcFace representation at a false acceptance rate (FAR) of 0.1\%, StyleGAN3 and DCFace have a capacity upper bound of $1.43\times10^6$ and $1.190\times10^4$, respectively; (b) the capacity reduces drastically as we lower the desired FAR with an estimate of $1.796\times10^4$ and $562$ at FAR of 1\% and 10\%, respectively, for StyleGAN3; (c) there is no discernible disparity in the capacity w.r.t gender; and (d) for some generative models, there is an appreciable disparity in the capacity w.r.t age. Code is available at \url{https://github.com/human-analysis/capacity-generative-face-models}.
\end{abstract}

%% file: 02-introduction.tex
\section{Introduction}

Generative face models have witnessed rapid progress and broad applicability in various practical applications: face image manipulation, animation, enhancement, synthetic face generation for training models for facial analysis tasks, and generating artistic faces, to name a few. The key desiderata for generative face models are 1) \emph{photorealism}, which refers to the resolution of generated faces and fidelity of facial details, 2) \emph{diversity}, which refers to the variations in the generated faces, in terms of pose, and facial attributes such as gender, age, ethnicity, and 3) \emph{uniqueness}, which refers to the distinctness of generated facial identities. Fueled by copious amounts of data, ever-growing computational resources, and algorithmic developments, current state-of-the-art generative models can create faces with 1) very high levels of photorealism at high-resolutions~\cite{karras2017progressive, karras2019style, karras2020analyzing, karras2021alias}, and 2) increasing levels of diversity~\cite{ghosh2018multi,hoang2018mgan}. Despite this tremendous progress, a crucial question about the uniqueness of generated faces still needs to be addressed, namely, \emph{what is the maximal number of unique identities can a given generative face model generate?} Answering this question is the central aim of this paper. Our contribution is to objectively determine the biometric capacity without generating face images at scale and conducting exhaustive empirical evaluations.

\begin{figure}
    \centering
    \begin{subfigure}{0.48\textwidth}
        \centering
        \includegraphics[trim=0 0.5cm 0 0, clip, width=\textwidth]{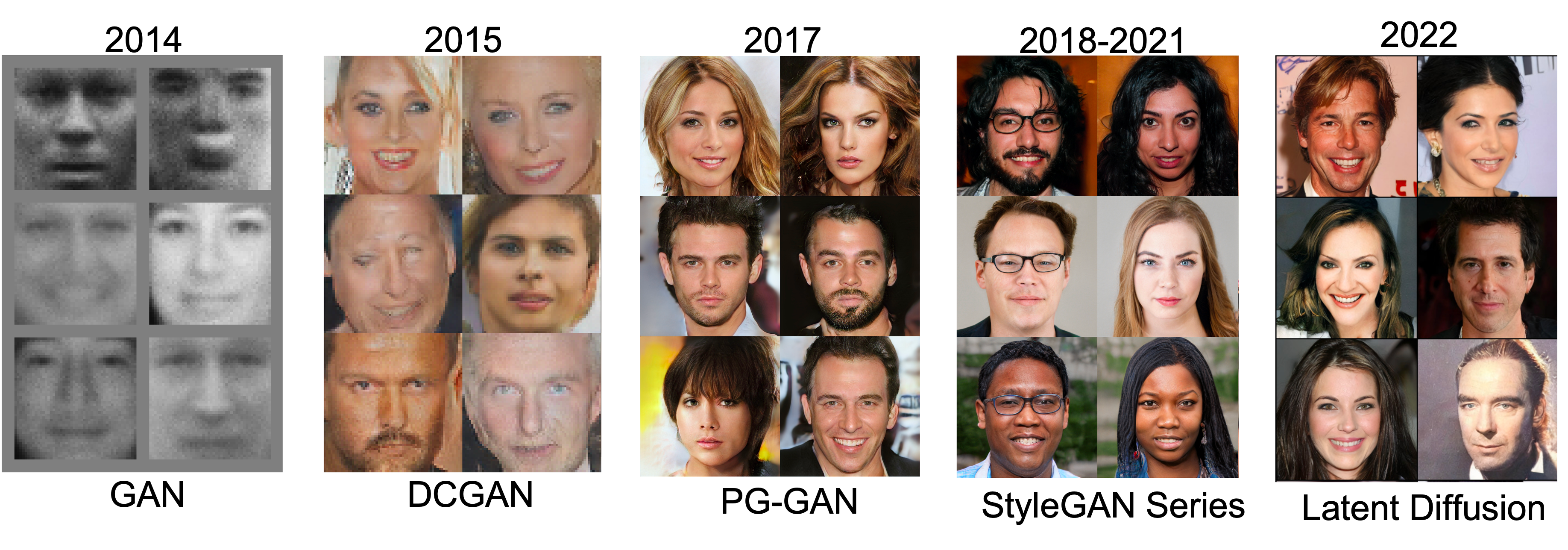}
        \caption{Photorealism\label{fig:photorealism}}
    \end{subfigure}
    \begin{subfigure}{0.235\textwidth}
        \centering
        \includegraphics[width=\textwidth]{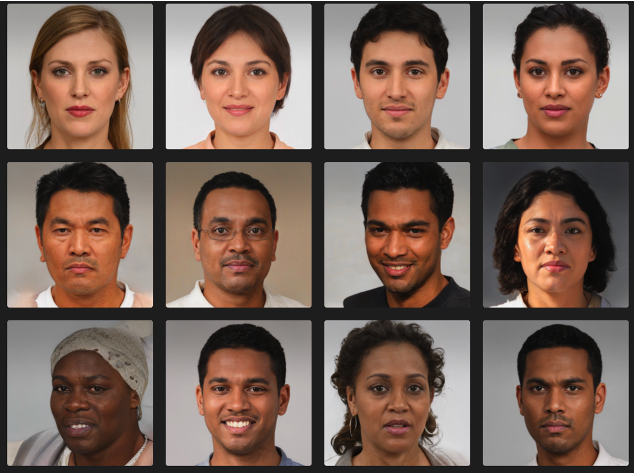}
        \caption{Diversity\label{fig:photorealism}}
    \end{subfigure}
    \begin{subfigure}{0.235\textwidth}
        \centering
        \includegraphics[width=\textwidth]{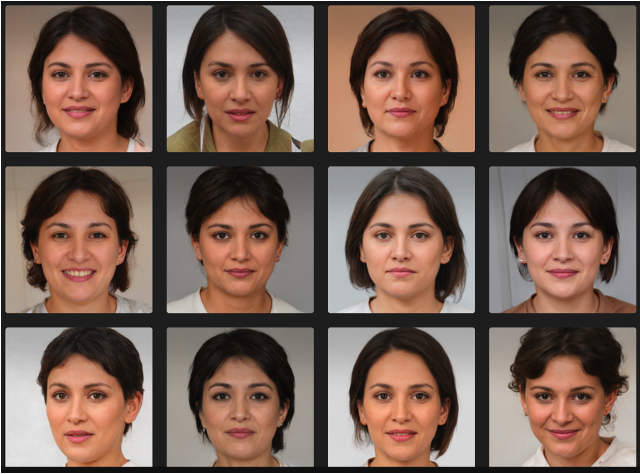}
        \caption{Uniqueness\label{fig:photorealism}}
    \end{subfigure}
    \caption{Three desiderata of generative face models, photorealism, diversity, and uniqueness. While there is significant work on improving photorealism and a few attempts at improving diversity, there is no study on the uniqueness of generated faces.\label{fig:teaser}}
\end{figure}

\begin{figure*}[!ht]
    \centering
    \includegraphics[width=\textwidth]{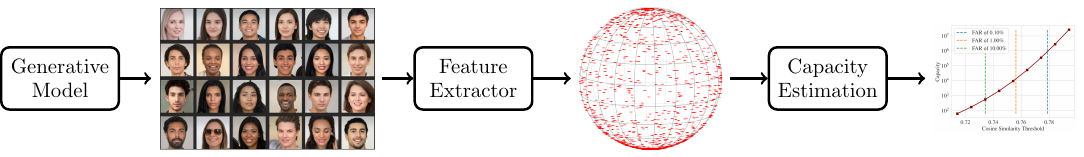}
    \caption{\textbf{Capacity Estimation Overview:} Given a generative face model, we first sample a collection of images. The images are then embedded in a hyperspherical space corresponding to a face feature extractor. A geometric model, in terms of the ratio of hyperspherical caps, is employed to estimate the capacity of the generative face models as a function of a desired match/no-match score threshold.\label{fig:overview}}
\end{figure*}
The ability to determine capacity affords the following benefits: 1) statistical estimates of the upper bound on the number of identities that generative face models can generate, which would allow for the informed deployment of such systems based on the expected scale of operation; 2) estimate the maximal scalability of a generative face model \emph{without} having to evaluate it at that scale exhaustively. Consequently, capacity offers an alternative yet crucial metric for comparing different generative face models in terms of the \emph{uniqueness} of generated facial identities.

An attractive solution for estimating the capacity of generative face models is to cast it as a packing problem\footnote{A generalization of the well-studied sphere-packing problem.}; the maximal number of shapes that can be fit, without overlapping, within the geometric support of the representation space. A loose bound on this packing problem can be obtained as a ratio of the support space's and shape's volumes. In the context of generative face models, the image support can be modeled as a low-dimensional population manifold $\mathcal{M}$ embedded within a high-dimensional image space $\mathcal{P}$. Similarly, images of each identity can also be modeled as its own manifold $\mathcal{M}_c \subseteq \mathcal{M}$. Under this setting, a bound on the capacity of the generative model can be obtained as a ratio of the volumes of the population and class-specific manifolds. 

Adopting the above approach to get empirical estimates of the capacity, however, does present a few challenges:
\begin{enumerate}
\item Estimating the support of the population manifold $\mathcal{M}$ and the class-specific manifolds $\mathcal{M}_c$, especially for high-dimensional data, such as face images, is an open problem.
\item Obtaining reliable estimates of the volume of arbitrarily shaped high-dimensional manifolds for the capacity bound is another open problem.
\end{enumerate}

In this paper, we propose a framework that addresses the challenges mentioned above to obtain reliable estimates of the capacity of any generative face model representation. Our solution relies on (1) modeling the generated faces in a low-dimensional face representation space that lies on the surface of a hypersphere, (2) approximating the population manifold by a hyperspherical cap in the feature space, (3) approximating the class-specific manifolds by a hyperspherical cap as a function of the specified false acceptance rate (FAR), and (4) estimating the capacity as a ratio of the surface area of the population and class-specific hyperspherical caps. Figure \ref{fig:overview} provides a pictorial illustration of the geometrical structure of our setting. The key contributions of this paper are:
\begin{enumerate}
    \item A robust statistical method for estimating an upper bound on the capacity of generative face models from a finite number of images sampled from such models.
    \item A geometrical model of capacity for faces embedded in a hyperspherical representation space as a function of a desired match/no-match similarity score threshold or false acceptance rate (FAR).
    \item The first practical attempt at estimating the capacity of generative face models. We consider both unconditional (PG-GAN~\cite{karras2017progressive}, StyleGAN family~\cite{kumari2022ensembling, karras2021alias}, Latent Diffusion Models~\cite{rombach2022high} and Generated Photos~\cite{generatedphotos}) and class-conditional generators (DCFace~\cite{kim2023dcface}).
    \item Studying the capacity of generative face models w.r.t demographic attributes such as gender and age.
\end{enumerate}

Numerical experiments suggest that our proposed approach can provide reasonable and reliable estimates of the upper bound of capacity, even as we vary the choice of the feature extraction space and the number of images/samples for estimating the capacity. Finally, among the generative face models we considered, most models do not show any disparity in the capacity w.r.t gender, while some models exhibit differences w.r.t age.

%% file: 03-related-work.tex
\section{Related Work}

\noindent\textbf{Generative Models for Faces:} Numerous generative face models have been proposed in the literature over the past decade. Recent models~\cite{brock2018large, choi2018stargan, ho2020denoising, karras2017progressive, karras2019style, karras2020analyzing, song2020denoising} allow high-fidelity generation of synthetic faces. Beyond generation, generative face models are also widely used to manipulate, animate and enhance face images~\cite{deng2020disentangled, hu2018disentangling, lin2018conditional, pumarola2018ganimation, sun2019single, tran2017disentangled, xiao2018elegant}. Such models operate by learning latent spaces that are disentangled w.r.t different face properties and controlling them selectively. More recently, latent variable models such as diffusion and score-based models~\cite{nichol2021improved, sohl2015deep, song2020denoising, song2021maximum, song2019generative, song2020improved} typify the state-of-the-art in generative face models. These models can be adapted for generating images conditioned on other information, such as text descriptions~\cite{ramesh2022hierarchical}. Collectively, these generative models represent the foundation of the dramatic improvements we are witnessing in the quality of generated faces and the broad range of emergent applications. However, despite this progress, there has been little to no effort in studying the \emph{uniqueness} of the generated identities regarding their maximal scalability. Addressing this gap is the primary goal of this paper.

\vspace{5pt}
\noindent\textbf{Capacity Estimation:} There have been efforts to determine the uniqueness of many biometric modalities, including fingerprint, iris, and face. Pankanti \etal \cite{pankanti2002individuality} and Zhu \etal \cite{zhu2007statistical} estimated the capacity of fingerprints by deriving an expression for estimating the probability of a false correspondence between minutiae-based representations from two arbitrary fingerprints belonging to two different fingers. Daugman~\cite{daugman2015information} proposed an information theoretic approach to compute the capacity of IrisCode. He first developed a generative model of IrisCode based on Hidden Markov Models and then estimated the capacity of IrisCode by calculating the entropy of this generative model. Gong~\etal~\cite{gong2017capacity} estimated the capacity of neural network-based face representation in high-dimensional Euclidean space. Finally, Terh{\"o}rst \etal~\cite{terhorst2022limited} proposed a theoretical model of face capacity, which was estimated through simulations in low dimensions and extrapolated to higher dimensions. While the work mentioned above focussed on estimating the capacity for real data and feature extractors, this paper focuses on estimating the capacity of generative face models.

%% file: 04-approach.tex
\section{Approach}
Given a generative face model, we first sample $N$ face images\footnote{We operate under the scenario where $N << C$, where $C$ is the capacity.} from the model. These images are represented in a vector space by extracting features from a face recognition model. Then, we employ our statistical capacity estimation model in the feature space and estimate capacity as a function of the corresponding face matcher's similarity score. This process is illustrated in Fig.\ref{fig:overview}.

\subsection{Image Representation}
The choice of image representation critically affects the biometric capacity estimates. We note that estimating the capacity directly from the pixel representations of the generated images is not desirable for several reasons. First, raw image pixels entangle identity and geometric and photometric variations. Moreover, since we aim to estimate capacity w.r.t. unique \emph{identities} instead of unique \emph{images}, we need to estimate capacity in a representation space that preserves identity while being invariant to other factors. Thus, a face recognition system's feature space is a well-justified representation choice. In summary, we posit that estimating a generator's \emph{biometric capacity} is inextricably linked to a biometric feature extractor.

We employ state-of-the-art face feature extractors (viz., AdaFace~\cite{kim2022adaface} and ArcFace~\cite{deng2019arcface}) and represent each image in the respective feature space. Such a design choice overcomes the challenges of using a pixel representation: 1) the features are robust to noise and other geometric (pose) and photometric (illumination) variations and instead comprise features related to biometric identity, and 2) the features lie on the surface of a sphere, and as we discuss next, capacity can be estimated reliably in this feature space.

\begin{figure}
    \centering
    \begin{subfigure}{0.23\textwidth}
        \centering
        \includegraphics[width=\textwidth]{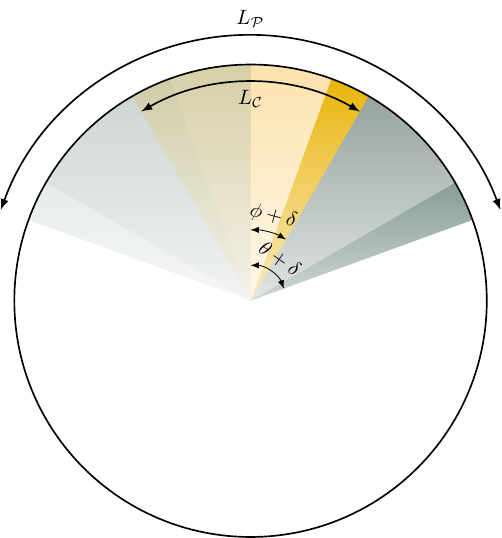}
        \caption{\label{fig:capacity-2d}}
    \end{subfigure}
    \begin{subfigure}{0.23\textwidth}
        \centering
        \includegraphics[width=\textwidth]{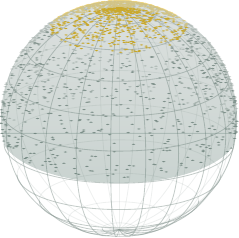}
        \caption{\label{fig:capacity-3d}}
    \end{subfigure}
    \caption{Illustration of capacity estimation concept for hyperspherical feature spaces. The population and identity span are shown in green and yellow, respectively. (a) In 2D, capacity is the ratio of the arc length of the population $\mathcal{P}$ and for an identity $\mathcal{C}$. (b) In 3D and higher dimensions, capacity is the ratio of the surface areas spanned by the population and single identity.\label{fig:capacity}}
\end{figure}

\subsection{Capacity Estimation Model\label{sec:capacity}}
\noindent\textbf{Concept:} We first provide a conceptual description of our capacity estimation approach by considering an illustrative 2D representation space. Under our choice of image representation, the features in this 2D space lie on a circle. The population of all faces spans an angle $2\theta$, while the face images from a single identity span an angle $2\phi$. Since we seek to estimate capacity as a function of the face matcher's false acceptance rate (FAR), we extend the population and identity-specific span by an angle $2\delta$, which varies with the desired FAR. Correspondingly, the angles spanned by the population and each identity are $2(\theta+\delta)$ and $2(\phi+\delta)$, respectively. The capacity can now be defined as the ratio of the arc lengths spanned by the population, $r\times2(\theta + \delta)$, and each identity, $r\times2(\phi + \delta)$, where $r$ is the radius of the circle. Therefore, the capacity $C(\theta, \phi, \delta) = \frac{\theta + \delta}{\phi + \delta}$. In 3D and higher dimensions, the features span the surface of a sphere instead of an arc. In this case, the capacity can be estimated as the ratio of the surface areas subtended by the solid angle $2(\theta+\delta)$ and $2(\phi+\delta)$. A pictorial illustration of this conceptual idea in 2D and 3D is shown in Fig.~\ref{fig:capacity}.

\vspace{3pt}
\noindent\textbf{General Case:} Let $\mathbb{S}^n$ be a $n$-hypersphere or a $n$-sphere for short, of radius $r$ in $n$-dimensional Euclidean space, i.e., 
\begin{equation}
    \mathbb{S}^n = \{\mathbf{x} \in \mathbb{R}^n:\|\mathbf{x}\|=r\}
\end{equation}

The area of the hypersphere $A_n(r)$ is,
\begin{equation}
    A_n(r) = \frac{2\pi^{n/2}}{\Gamma\left(\frac{n}{2}\right)}r^{n-1}
\end{equation}
\noindent where, $\Gamma$ is the gamma function. Let a hyperspherical cap subtend a solid angle of $2\Omega$, where $0\leq\Omega\leq\pi/2$. The area $A^{\Omega}_n(r)$ of such a hyperspherical cap is,
\begin{equation}
    A^{\Omega}_n(r) = \frac{1}{2}A_n(r)I_{sin^2(\Omega)}\left(\frac{n-1}{2},\frac{1}{2}\right)
\end{equation}
\noindent where, $I_x(a,b)$ is the regularized incomplete beta function. Given two hyperspherical caps with solid angle $2\Omega_1=2(\theta+\delta)$ and $2\Omega_2=(\phi+\delta)$, their ratio, which corresponds to our capacity estimate, is given by
\begin{equation}
    C(\theta, \phi, \delta) = \frac{I_{sin^2(\Omega_1)}\left(\frac{n-1}{2},\frac{1}{2}\right)}{I_{sin^2(\Omega_2)}\left(\frac{n-1}{2},\frac{1}{2}\right)}
    \label{eq:capacity}
\end{equation}

Note that we only have access to $cos(\theta)$, $\cos(\phi)$, and $cos(\delta)$ from the cosine similarity scores between the face features. The value of $sin^2(\Omega_1)$ can be estimated as,
\begin{equation}
   \begin{aligned}
    sin^2(\Omega_1) &= 1 - cos^2(\Omega_1) \mbox{, where,}\\
    cos(\Omega_1) &= cos(\theta)cos(\delta) - sin(\theta)sin(\delta)
   \end{aligned}
\end{equation}
Similarly, the value of $sin^2(\Omega_2)$ can be estimated as,
\begin{equation}
   \begin{aligned}
    sin^2(\Omega_2) &= 1 - cos^2(\Omega_2) \mbox{, where,}\\
    cos(\Omega_2) &= cos(\phi)cos(\delta) - sin(\phi)sin(\delta)
   \end{aligned}
\end{equation}

Given $\Omega_1$ and $\Omega_2$, the capacity from \eqref{eq:capacity} is exact without approximations. Accordingly, our capacity estimates are reliable to the extent that the estimates of $\theta$ and $\phi$ are reliable.

%% file: 05-experiments.tex
\begin{figure*}[!ht]
    \centering
    \begin{subfigure}{0.16\textwidth}
        \centering
        \includegraphics[width=\textwidth]{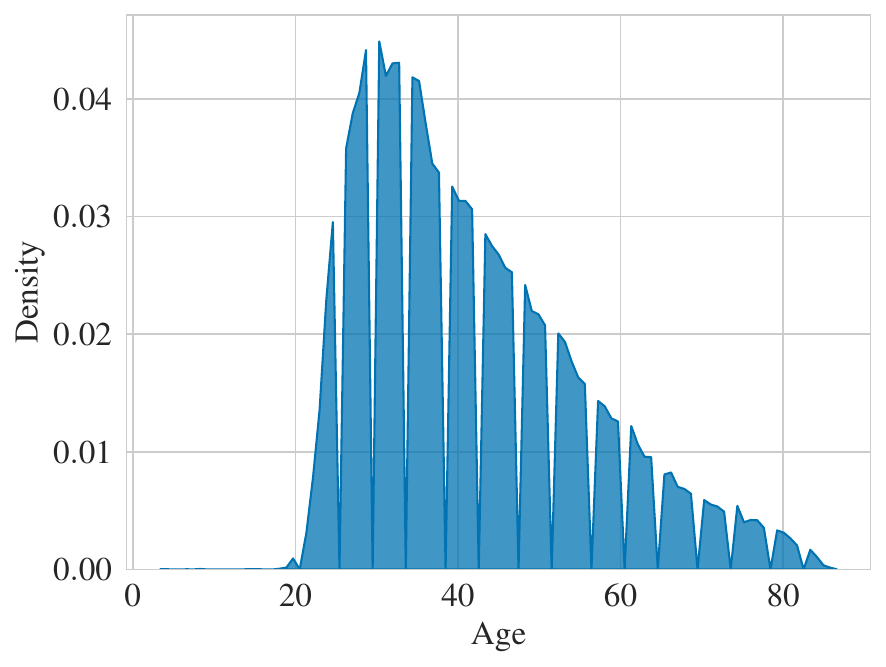}
        \includegraphics[width=\textwidth]{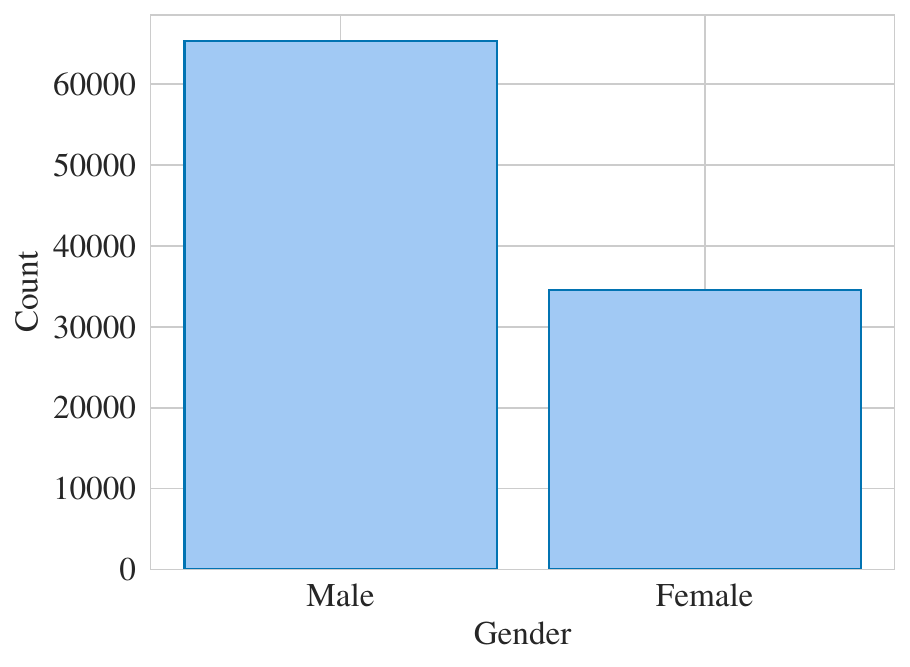}
        \caption{PG-GAN\label{fig:stats-pg-gan}}
    \end{subfigure}
    \begin{subfigure}{0.16\textwidth}
        \centering
        \includegraphics[width=\textwidth]{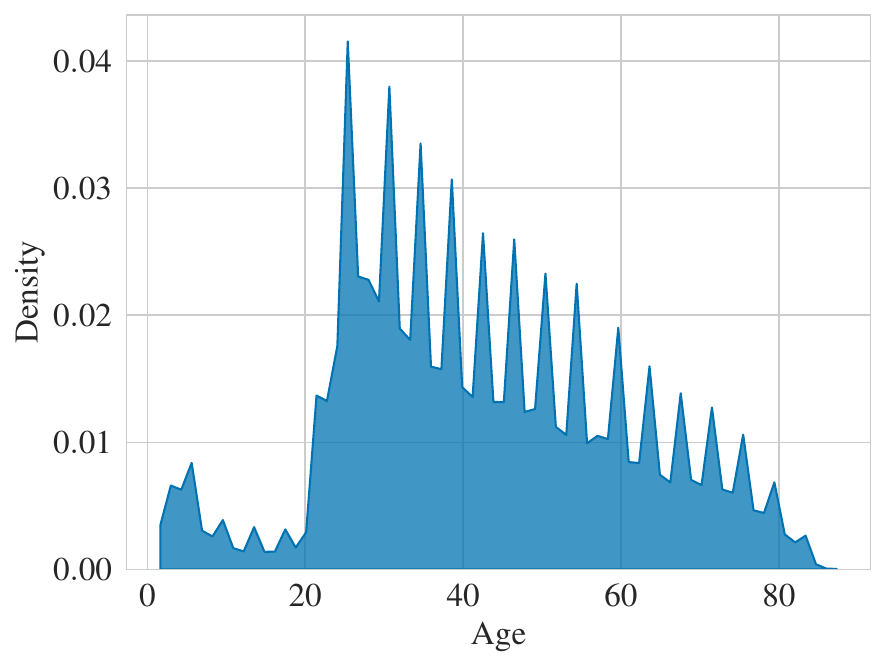}
        \includegraphics[width=\textwidth]{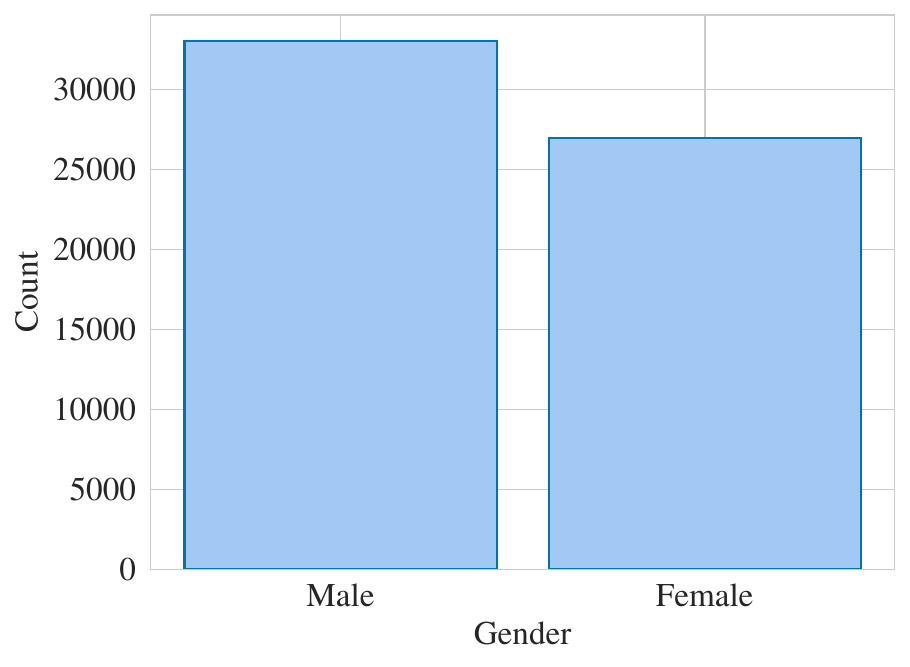}
        \caption{StyleGAN2+ADA\label{fig:stats-stylegan2}}
    \end{subfigure}
    \begin{subfigure}{0.16\textwidth}
        \centering
        \includegraphics[width=\textwidth]{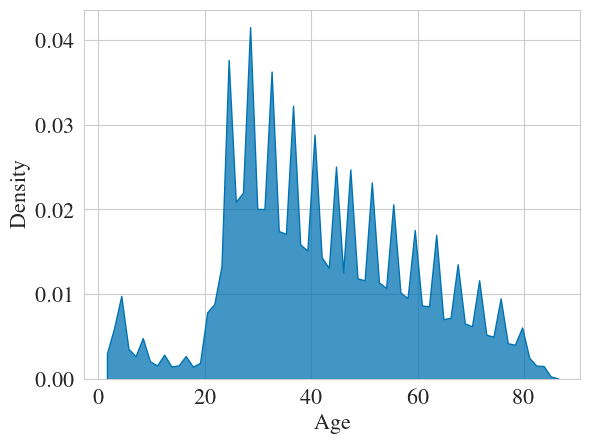}
        \includegraphics[width=\textwidth]{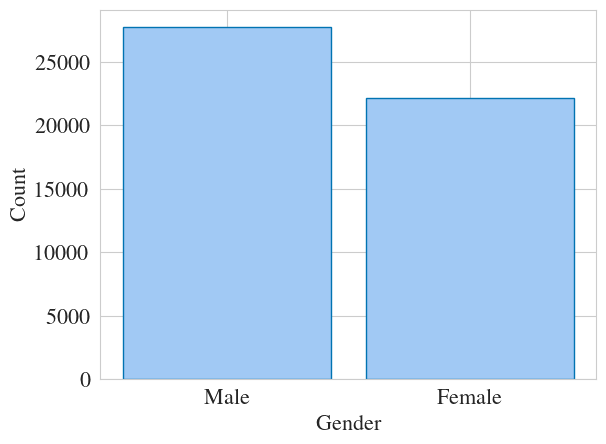}
        \caption{StyleGAN3\label{fig:stats-stylegan3}}
    \end{subfigure}
    \begin{subfigure}{0.16\textwidth}
        \centering
        \includegraphics[width=\textwidth]{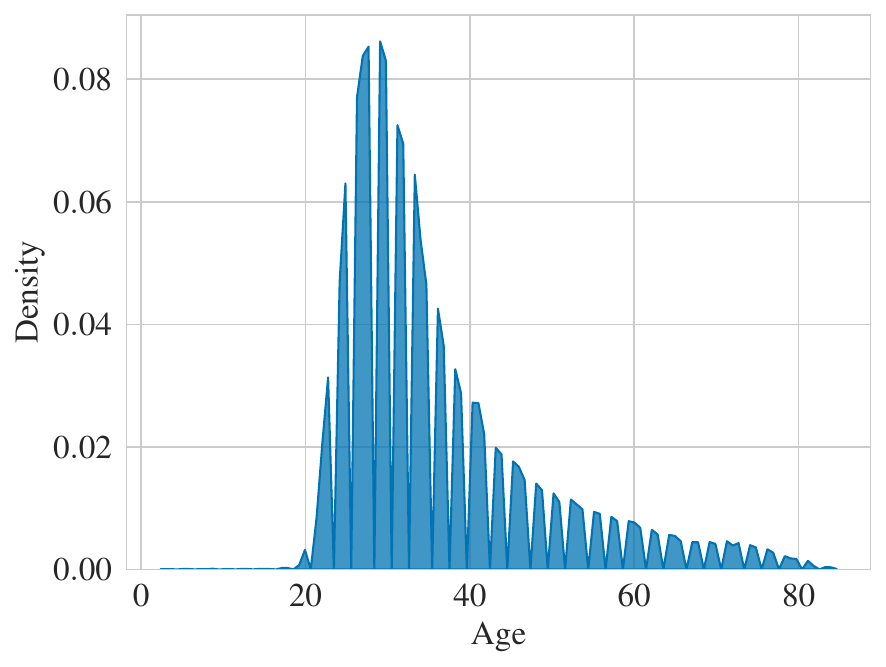}
        \includegraphics[width=\textwidth]{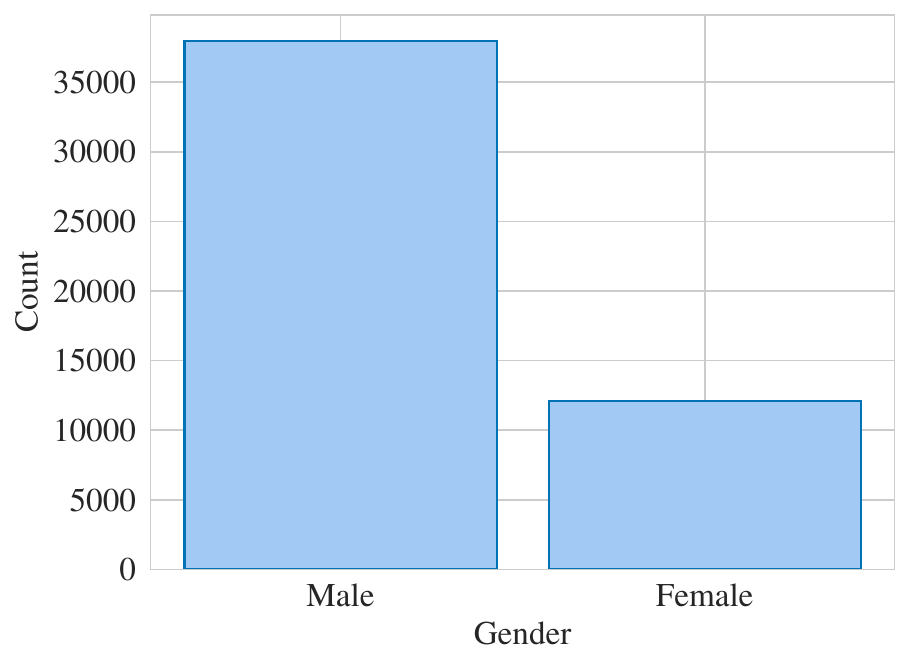}
        \caption{LDM\label{fig:stats-ldm}}
    \end{subfigure}
    \begin{subfigure}{0.16\textwidth}
        \centering
        \includegraphics[width=\textwidth]{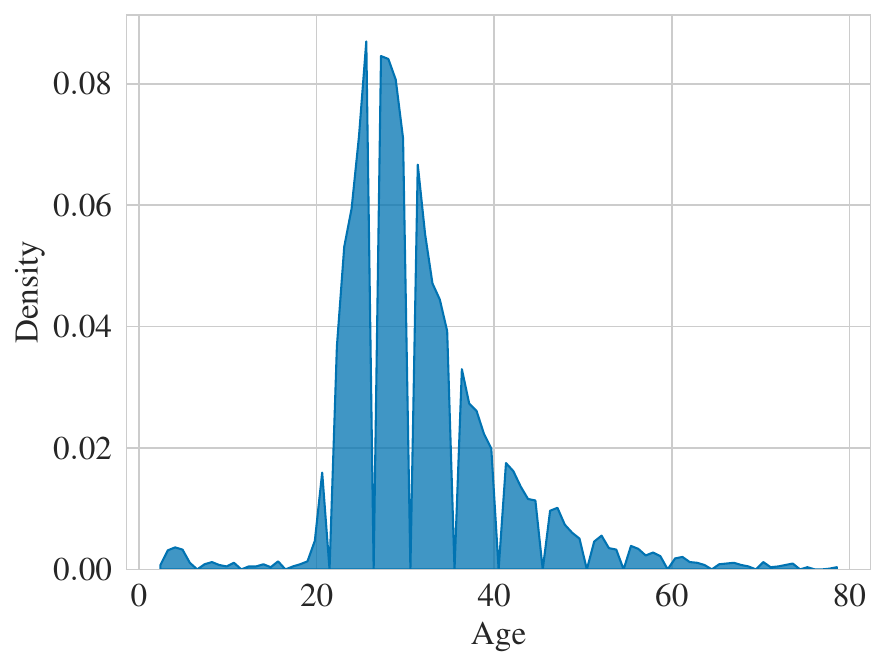}
        \includegraphics[width=\textwidth]{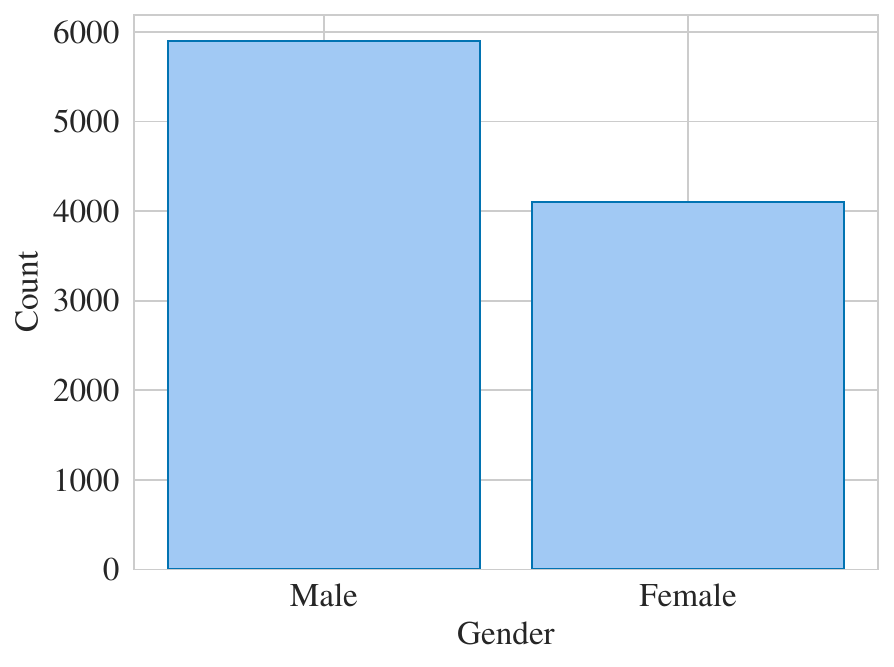}
        \caption{Generated Photos\label{fig:stats-generated-photos}}
    \end{subfigure}
    \begin{subfigure}{0.16\textwidth}
        \centering
        \includegraphics[width=\textwidth]{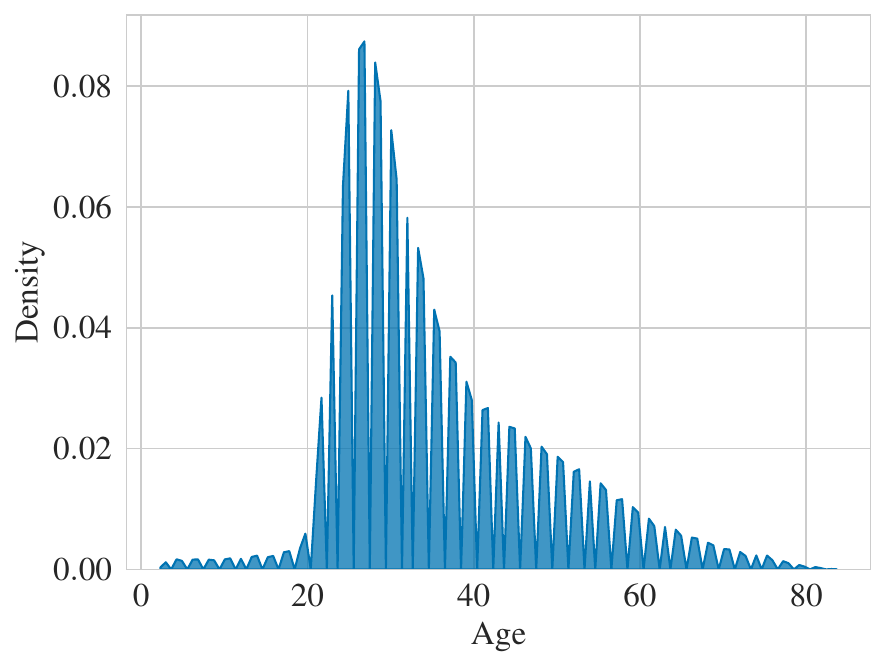}
        \includegraphics[width=\textwidth]{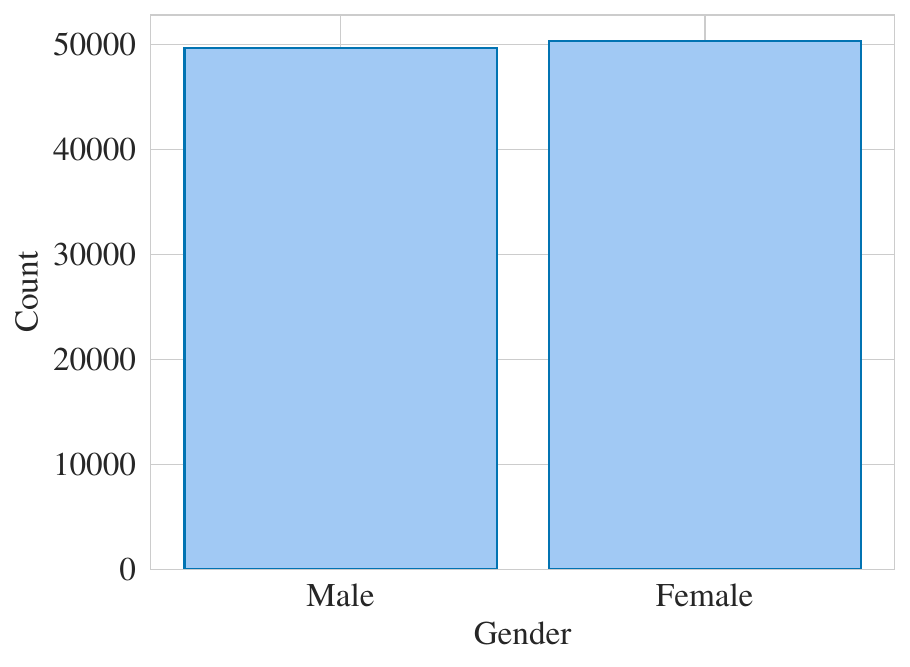}
        \caption{DCFace (CC)\label{fig:stats-dcface}}
    \end{subfigure}
    \caption{Age (top) and gender (bottom) statistics estimated using ArcFace for the images generated by different generative face models.\label{fig:dataset-stats}}
\end{figure*}

\section{Experiments}
We will estimate the capacity of multiple generative models by employing the above theoretical capacity model with multiple feature extractors. Then, we will study 1) capacity across different demographic groups and 2) the impact of design choices on the stability of capacity estimates.

\subsection{Generative Models}
We consider six generative models, including five unconditional (four open-source and one commercial) and one class-conditional (CC) generator, spanning a range of training datasets and model architecture combinations.

\vspace{3pt}
\noindent\textbf{PG-GAN~\cite{karras2017progressive}:} This was the first generative model designed specifically for generating high-resolution images. The authors also introduced the CelebA-HQ image dataset, a high-resolution subset of the CelebA dataset with 30,000 images. PG-GAN introduced an architecture that progressively increases the image's resolution to $1024\times 1024$. We used 50,000 images generated from PG-GAN.

\vspace{3pt}
\noindent\textbf{StyleGAN2~\cite{karras2019style}+ADA~\cite{ADA}:} A GAN-based architecture designed for explicit control over the style of the generated images and to learn from a limited number of training samples using an ensemble of discriminators~\cite{kumari2022ensembling}. The model was trained on the FFHQ dataset with images at a resolution of $1024\times 1024$. We used 50,000 images generated from StyleGAN2+ADA.

\vspace{3pt}
\noindent\textbf{StyleGAN3~\cite{karras2021alias}:} The most recent version of the StyleGAN family of generative models was designed to mitigate the effects of aliasing in the generator architecture. While the photorealism of the images is similar to those from StyleGAN2~\cite{karras2020analyzing}, the internal representations and the learned latent space were fully equivariant to translation and rotation even at subpixel scales. We used 50,000 images from the model trained on the FFHQ dataset.

\vspace{3pt}
\noindent\textbf{Latent Diffusion Model~\cite{rombach2022high} (LDM):} A generative model that includes an autoencoder and learns a denoising diffusion probabilistic model in the autoencoder's latent space. We consider the LDM version trained on the CelebA-HQ dataset at a resolution of $256\times 256$. We used 50,000 images generated from LDM.

\vspace{3pt}
\noindent\textbf{Generated Photos~\cite{generatedphotos}:} A commercial website that offers synthetic images generated by a StyleGAN-based model trained on a proprietary and curated dataset of images of models. The training face images were captured in a photo studio under controlled lighting conditions and with similar variations. The studio subjects were carefully selected to span demographic attributes like gender and ethnicity. After generation, further processing was employed to remove the background in the images. We used the academic version of this dataset comprising 10,000 generated face images.

\vspace{3pt}
\noindent\textbf{DCFace~\cite{kim2023dcface}:} Dual Condition Face Generator is a diffusion-based face generator. It is designed to explicitly control inter-class and intra-class variations of faces through explicit control of the subject's appearance (ID) and external factors (style). Synthetic faces generated from this dataset were used for training a face recognition system. The model was trained on the FFHQ dataset. Unlike the other generative face models we consider, DCFace is a class-conditional (CC) face generator that affords explicit control over the identity of the generated images. Thus, unlike the case with the above-mentioned unconditional generators, the span of a single identity $\phi$ can be directly estimated for DCFace. We used 100K images generated from DCFace with 10,000 identities and ten images per identity.

\subsection{Face Feature Extractors\label{subsec:face-feature-extractors}}
We employ two face feature extractors that are the state-of-the-art models commonly used in face verification and recognition tasks: \textbf{ArcFace}~\cite{deng2019arcface} trained on WebFace600K~\cite{zhu2021webface260m} and \textbf{AdaFace}~\cite{kim2022adaface} trained on MS1MV2~\cite{guo2016ms,deng2019arcface}. Both feature extractors map face images to a 512-dimensional hyperspherical feature space, i.e., $n=512$. ArcFace is trained for open-set face recognition by maximizing the geodesic distance between features. In contrast, AdaFace is trained with an adaptive margin loss function considering image quality. Both models exhibit similar face verification accuracy for high-quality images, which is the case with the generative images we consider in this paper. Therefore, we do not expect the capacity estimates to depend significantly on the choice of the image representation space between the two feature spaces.

\subsection{Statistics of Generated Face Images}
Besides high photorealism, faces generated by an ideal generative model must be diverse and unique. Therefore, we calculate the age and gender statistics of each generated dataset. Since true labels for these attributes are unknown, we estimate these statistics using ArcFace's predictions. 

Figure~\ref{fig:dataset-stats} (top row) shows the age statistics across the different generative face models. We make the following observations; 1) the most prominent age group in the datasets is 20-40, and 2) StyleGAN models, which have been trained on the FFHQ dataset, can generate faces of young individuals (ages 0-20) due to the explicit care taken during the dataset curation to include images with a large diversity in age. Therefore, we expect StyleGAN2+ADA and StyleGAN3 to exhibit only a slight disparity in capacity across the age groups. Analogously, the other models are expected to show significant bias towards age groups 20-40.

Figure~\ref{fig:dataset-stats} (bottom row) shows the gender statistics across the different generative face models. Except for LDM and PGGAN, we observe that most datasets are reasonably balanced w.r.t. gender, with each having slightly more images of males than females. Hence, we expect the capacity of the former two models to exhibit some disparity in the capacity across gender.

\subsection{Estimating population and intra-class variance\label{sec:intra-class-variance}}
As alluded to in Section~\ref{sec:capacity}, the reliability of our capacity estimates is critically dependent on the estimates of the population variance, $\theta$, and the intra-class variance, $\phi$.

\begin{figure*}
    \centering
    \subcaptionbox{PG-GAN}{\includegraphics[width=0.16\textwidth]{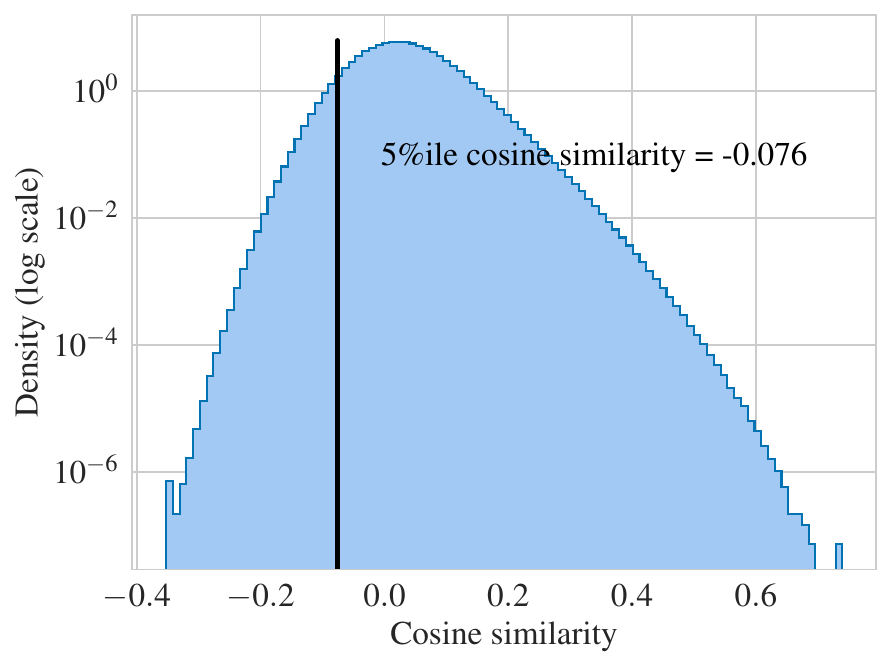}}
    \subcaptionbox{StyleGAN2+ADA}{\includegraphics[width=0.16\textwidth]{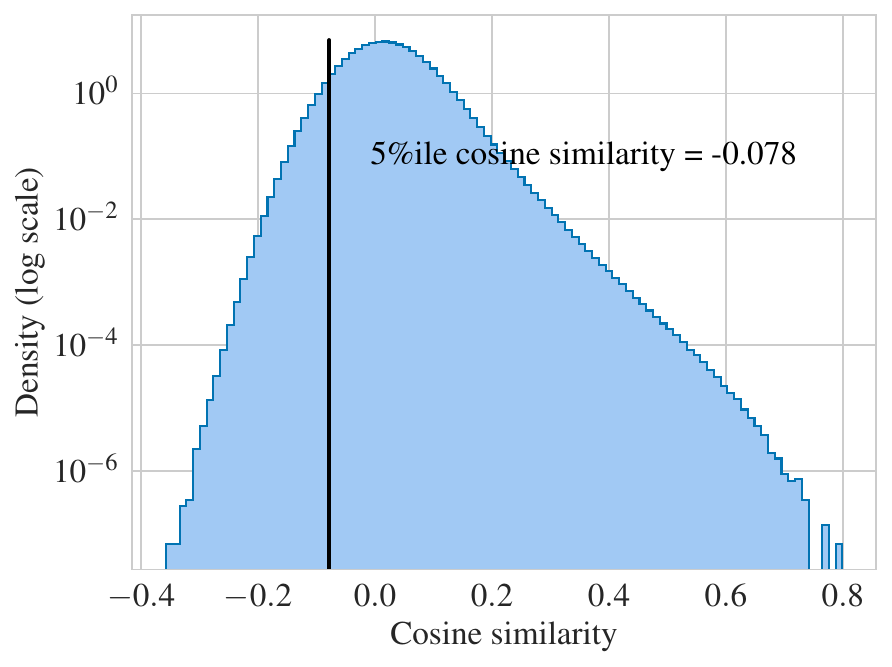}}
    \subcaptionbox{StyleGAN3}{\includegraphics[width=0.16\textwidth]{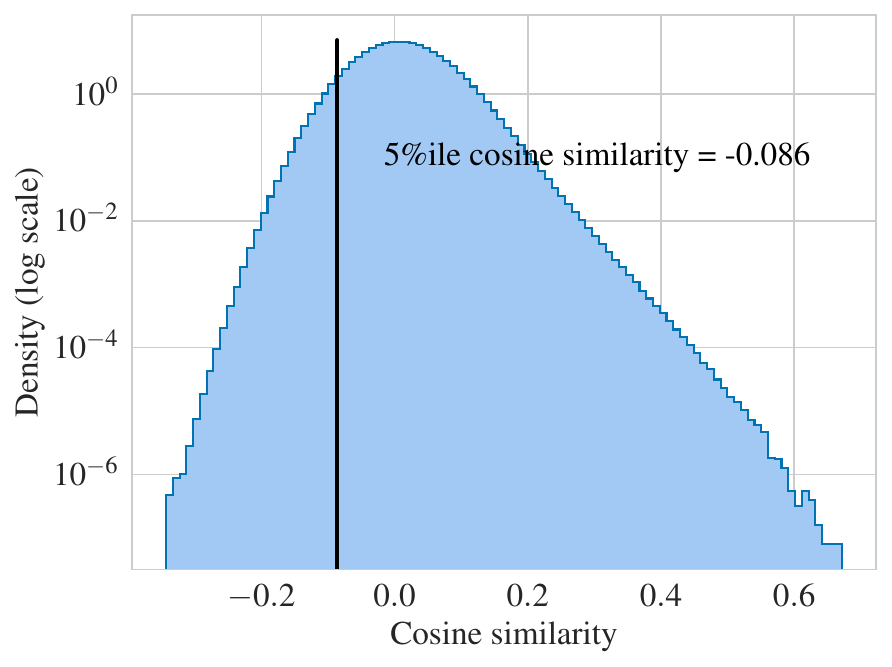}}
    \subcaptionbox{LDM}{\includegraphics[width=0.16\textwidth]{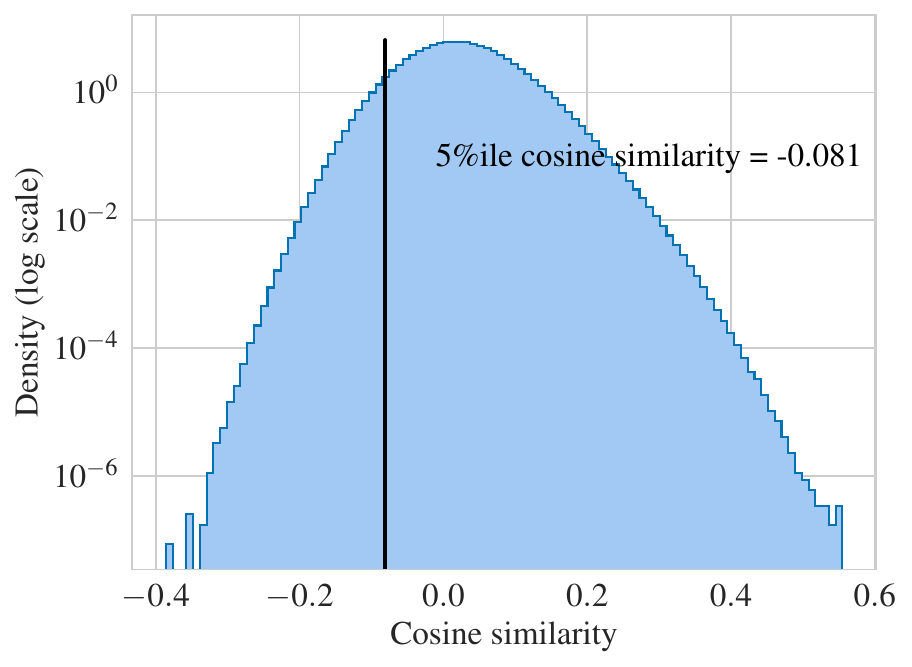}}
    \subcaptionbox{Generated Photos}{\includegraphics[width=0.16\textwidth]{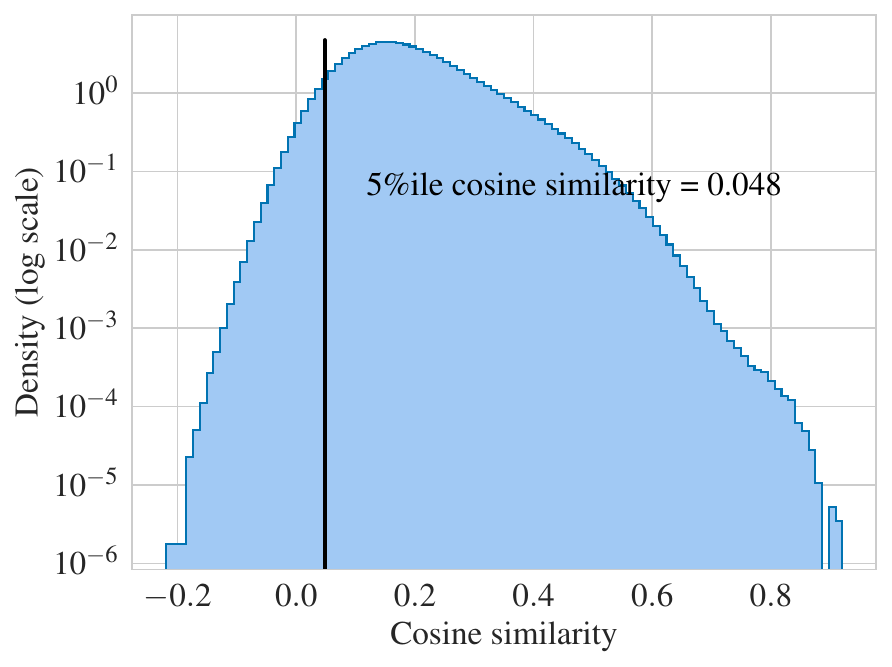}}
    \subcaptionbox{DCFace (CC)}{\includegraphics[width=0.16\textwidth]{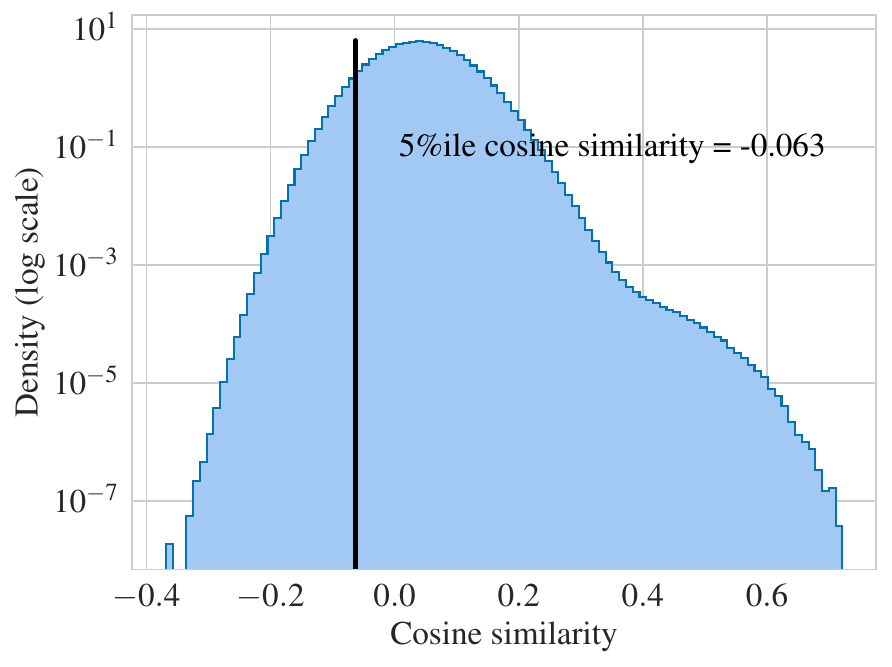}}
    \caption{Cosine similarity score (log scale) distribution across different generative models. The solid vertical line corresponds to the threshold value for the 5th percentile, which determines the population-level variance.\label{fig:similarity-score-distribution}}
\end{figure*}

The \textbf{population variance}, $\theta$, can be estimated by considering the range of cosine similarity scores spanned by all the images from a given generative face model. We compute the distance between all pairs of images in the feature space for each dataset and identify the furthest images from the score distribution. Furthermore, we consider the 5th percentile distance in the score distribution as the population variance for the respective datasets to account for outliers such as extremely poor-quality faces. Figure~\ref{fig:similarity-score-distribution} shows the similarity score distribution (log-scale) along with the score threshold, $s_{th}$, used to determine $\theta=\frac{\cos^{-1}(s_{th})}{2}$.

\begin{figure}
    \vspace{-0.25cm}
    \centering
    \includegraphics[width=0.48\textwidth]{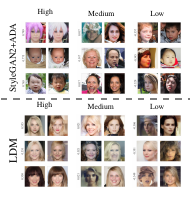}
    \caption{Image pairs with high, medium, and low similarity among the 50,000 faces generated by StyleGAN2+ADA~\cite{kumari2022ensembling} and LDM~\cite{rombach2022high}. The similarity scores are shown next to the corresponding image pair.\label{fig:similar-dissimilar-faces}}
    \vspace{-0.25cm}
\end{figure}

Ideally, \textbf{intra-class variance}, $\phi$, should be estimated from the distribution of similarity scores between images of a single identity. However, doing so for the unconditional face generators is challenging for two reasons. Firstly, manual labeling is necessary to identify images of the same class, which is time-consuming and error-prone. Secondly, unlike class-conditional models like DCFace, unconditional generative models do not afford explicitly controllable generation of images from the same identity. As such, we may be unable to generate multiple images of the same identity without sampling a huge number of images from such generative models. Examples of image pairs shown in Fig.~\ref{fig:similar-dissimilar-faces} illustrate these challenges. For instance, even images with the highest similarity score for StyleGAN2+ADA and LDM do not appear to be of the same identity.

To overcome this challenge, we assume that a representative value of the intra-class variance for the unconditional generative models can be estimated from the intra-class variance of the face matcher corresponding to the feature extractor we employ on a reference dataset of real-faces (LFW~\cite{huang2008labeled} in our experiments). Therefore, we select the threshold value of 0.2125 corresponding to a FAR of 0.1\%, which is strict enough to mimic the intra-class variance, i.e., $\cos(2\phi)$ of faces of a canonical identity. For DCFace, we first estimate the inter-class variance of each identity and select the median value (0.123 in ArcFace space).

\subsection{Capacity of Generative Face Models}
\begin{figure}
    \centering
    \includegraphics[width=0.48\textwidth]{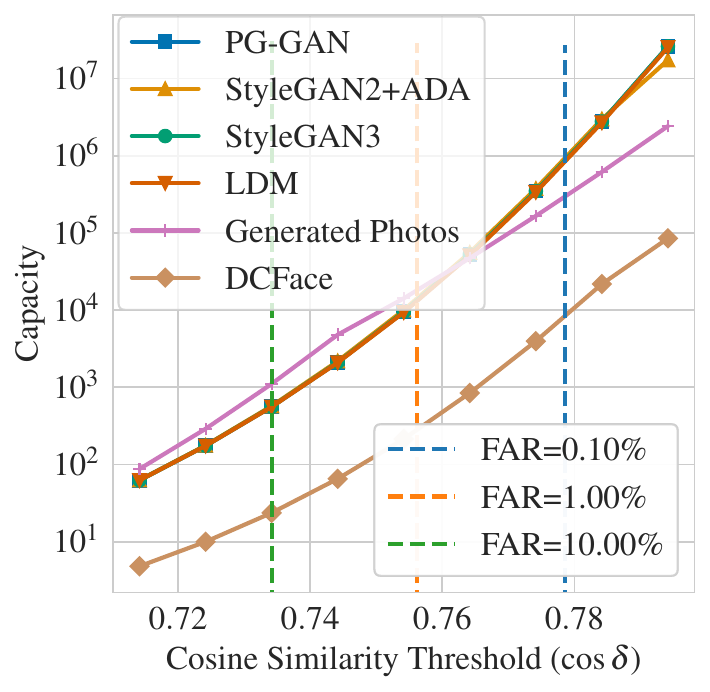}
    \caption{Comparison of capacity estimates across different generative models in the ArcFace feature space. Exact capacity values can be found at the \href{https://github.com/human-analysis/capacity-generative-face-models}{project page}.\label{fig:arcface-datasets}}
\end{figure}

Figure~\ref{fig:arcface-datasets} shows the capacity of all the generative face models we considered in this paper as a function of the cosine similarity threshold, $cos(\delta)$. \emph{We note that these capacity estimates are upper bounds of the actual capacity of the generative models.} The plots also show the threshold values corresponding to FAR values of 0.1\%, 1\%, and 10\% for the ArcFace-WebFace600K model. Although the indicated threshold values are for the LFW dataset, we study the effect of this choice in Section~\ref{subsec:ref-dataset-ablation}. We make the following observations from the results: 1) Most of the generative models, except ``Generated Photos" and DCFace, exhibit very similar capacity values as a function of the cosine similarity. As expected, the capacity increases with the cosine similarity threshold. 2) The capacity of the ``Generated Photos" dataset, which is curated to remove poor-quality images, is lower by one order of magnitude. This can be explained by the fact that ``Generated Photos" has a lower population variance (notice that the lower end of the score distribution in Fig.~\ref{fig:similarity-score-distribution} is higher than other models). 3) The capacity of DCFace, a class-conditional model, is lower by two orders of magnitude. This can be explained by the larger intra-class variance of $\arccos(0.123)$ for DCFace vs. $\arccos(0.2125)$ (threshold at FAR of 0.1\% for unconditional generators) for StyleGAN3. 4) While the capacity of ``Generated Photos" grows exponentially with the cosine similarity threshold, the capacity of the other generative models grows super-exponentially. 5) At higher FAR thresholds, ``Generated Photos" has a greater capacity than the other generative models, indicating that ``Generated Photos" generates more distinct images, i.e., in comparison, the other models generate more similar identities.
\begin{figure*}[!ht]
    \centering
    \includegraphics[width=0.95\textwidth]{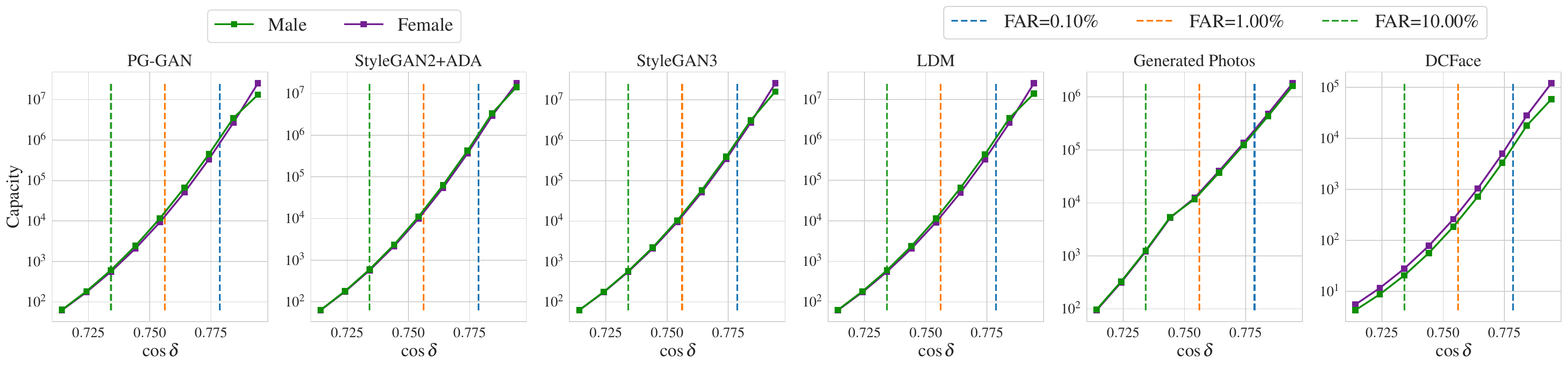}
    \caption{Capacity of generative models in terms of their ability to generate unique faces of different genders.\label{fig:capacity-arcface-gender}}
    \vspace{-0.25cm}
\end{figure*}
\begin{figure*}[!ht]
    \centering
    \includegraphics[width=0.95\textwidth]{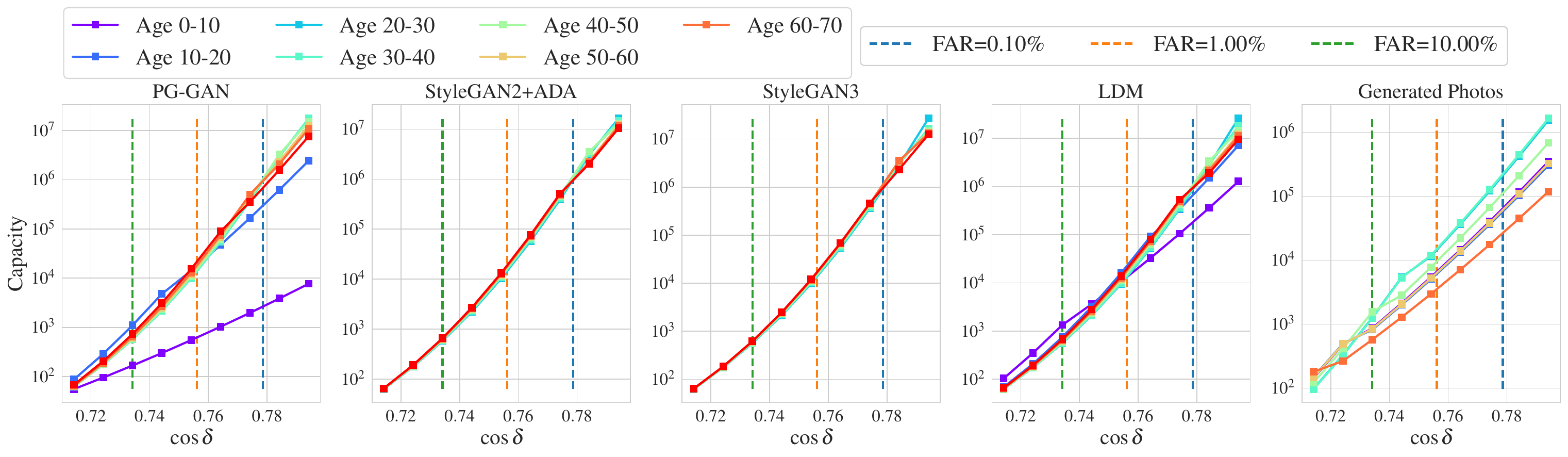}
    \caption{Capacity of generative models in terms of their ability to generate unique faces at different age groups.\label{fig:capacity-arcface-age}}
    \vspace{-0.25cm}
\end{figure*}

\subsection{Capacity Across Demographic Attributes}
We study the capacity of the generative face models for different demographic attributes such as gender and age. In the ideal case, generative models should be able to generate a similar number of identities across \emph{all} demographic labels. Figure~\ref{fig:capacity-arcface-gender} shows the capacity w.r.t gender\footnote{In this work, we only considered two genders.}. We observe that none of the models exhibit noticeable disparity in the capacity estimate for gender across all cosine similarity thresholds. This is due to the good coverage of males and females in the CelebA-HQ and FFHQ datasets on which the generative models were trained.

Figure~\ref{fig:capacity-arcface-age} shows capacity w.r.t age groups. Unlike the case with gender, here we observe a discernable disparity in capacity across age groups, especially for ``Generated Photos", PG-GAN, and LDM. Both PG-GAN and LDM are trained on CelebA-HQ, which does not have many images of faces under age 20. As a result, the age distribution in the generated images (see Figs.~\ref{fig:stats-pg-gan} and~\ref{fig:stats-ldm}) and the capacity for that age group are lower than the other age groups. At the same time, ``Generated Photos" is trained on curated data collected in a studio and includes a few but not many subjects under the age of 20 (see Fig.~\ref{fig:stats-generated-photos}). StyleGAN2+ADA and StyleGAN3, on the other hand, have been trained on the FFHQ dataset, which was deliberately constructed to have diverse images spanning different age groups. This is also apparent in the age distribution (see Fig.~\ref{fig:stats-stylegan2} and~\ref{fig:stats-stylegan3}) of the images generated from these models.
\begin{figure}[!ht]
    \centering
    \includegraphics[width=0.5\textwidth]{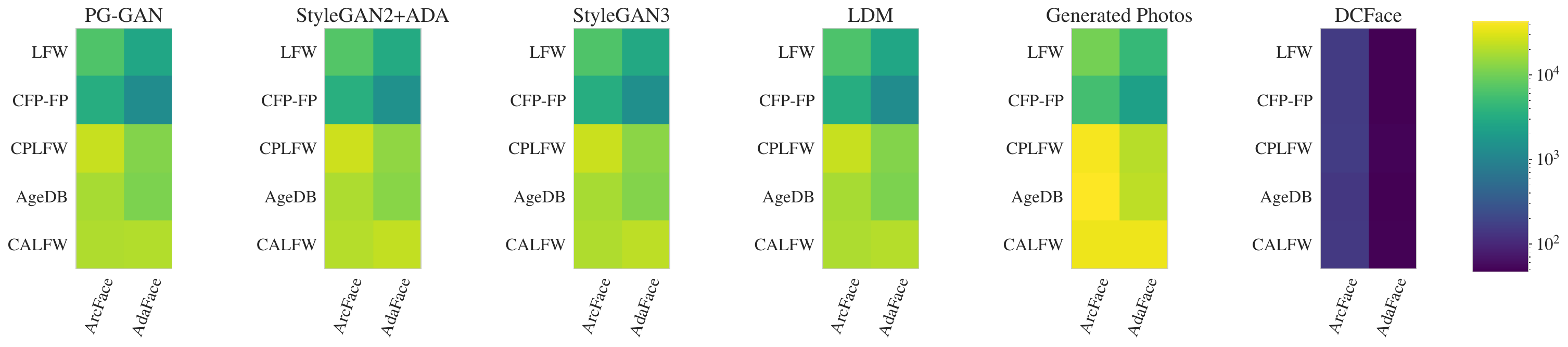}
    \caption{Effect of reference dataset for determining intra-class variance on the capacity estimate.\label{fig:ref-dataset-face-model-ablation}}
\end{figure}
\begin{figure*}[!ht]
    \centering
    \begin{subfigure}{0.16\textwidth}
        \centering
        \includegraphics[width=\textwidth]{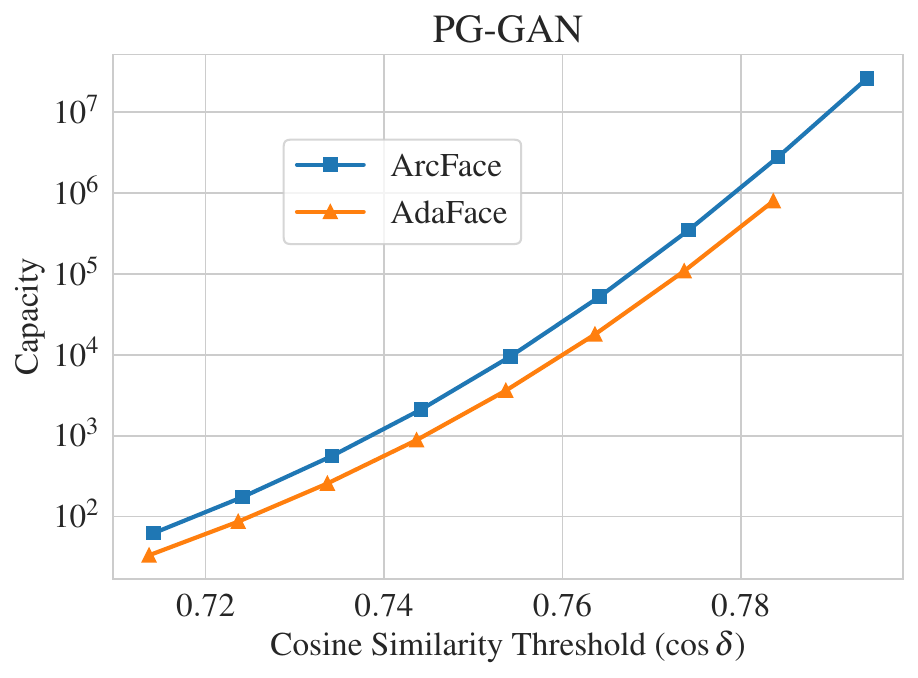}
    \end{subfigure}
    \begin{subfigure}{0.16\textwidth}
        \centering
        \includegraphics[width=\textwidth]{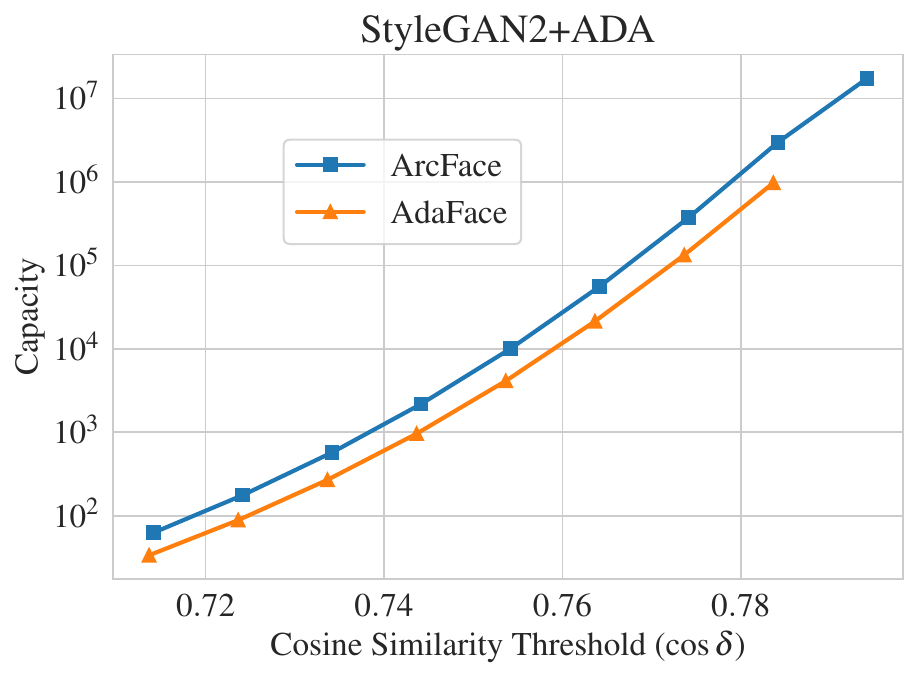}
    \end{subfigure}
    \begin{subfigure}{0.16\textwidth}
        \centering
        \includegraphics[width=\textwidth]{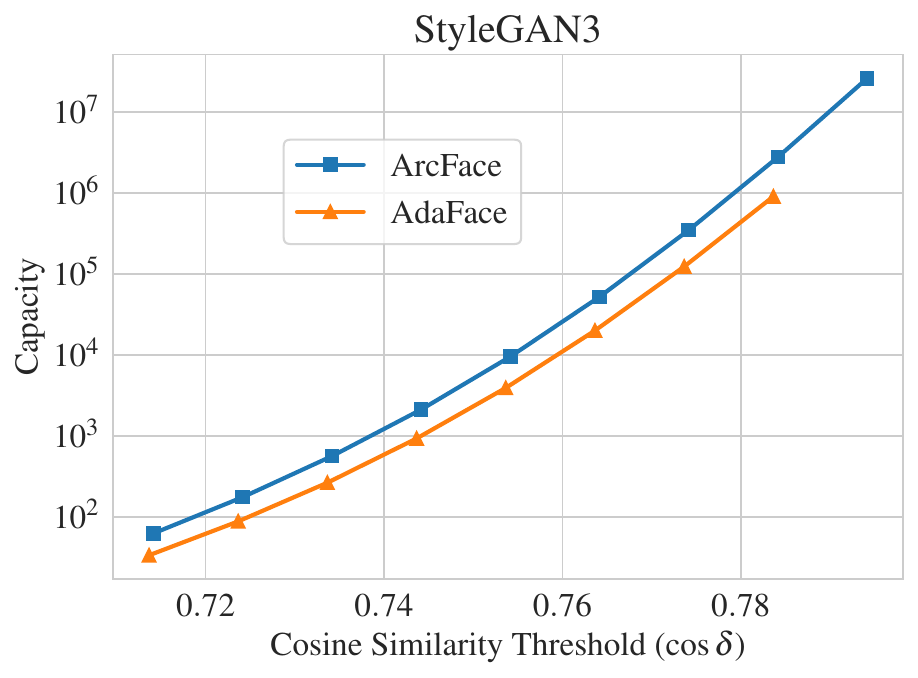}
    \end{subfigure}
    \begin{subfigure}{0.16\textwidth}
        \centering
        \includegraphics[width=\textwidth]{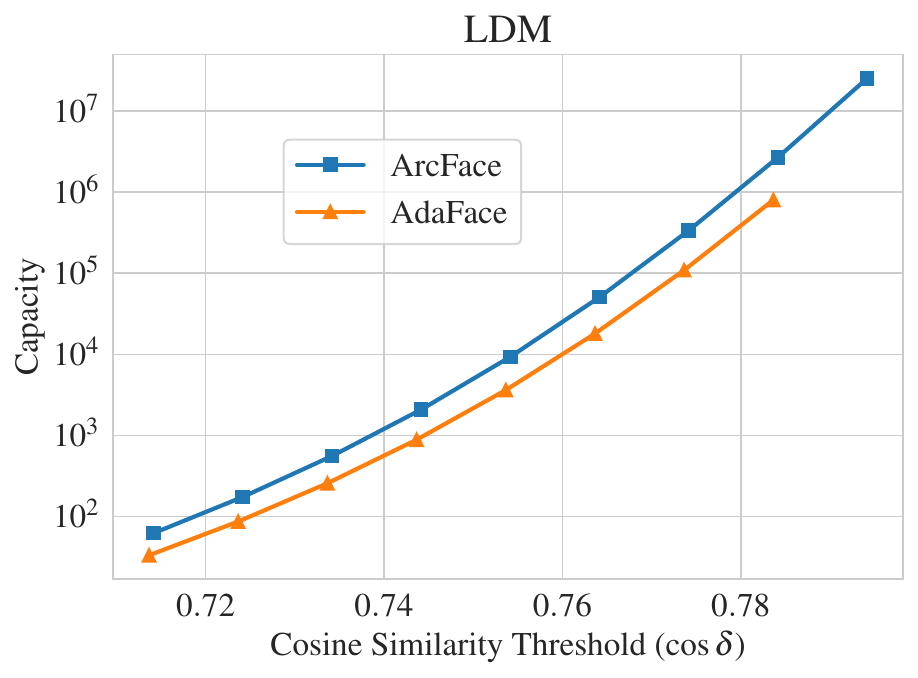}
    \end{subfigure}
    \begin{subfigure}{0.16\textwidth}
        \centering
        \includegraphics[width=\textwidth]{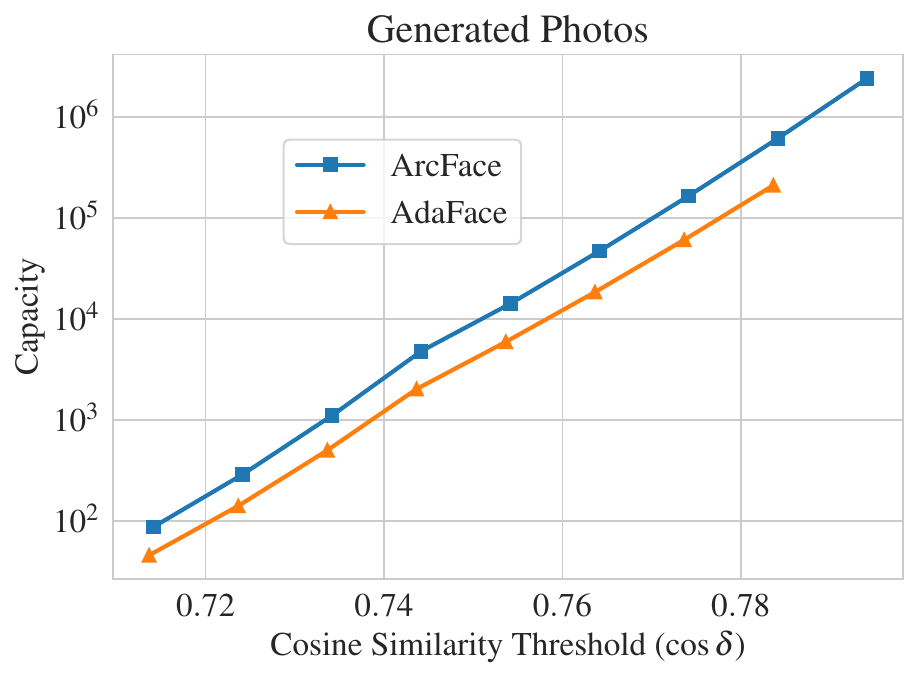}
    \end{subfigure}
    \begin{subfigure}{0.16\textwidth}
        \centering
        \includegraphics[width=\textwidth]{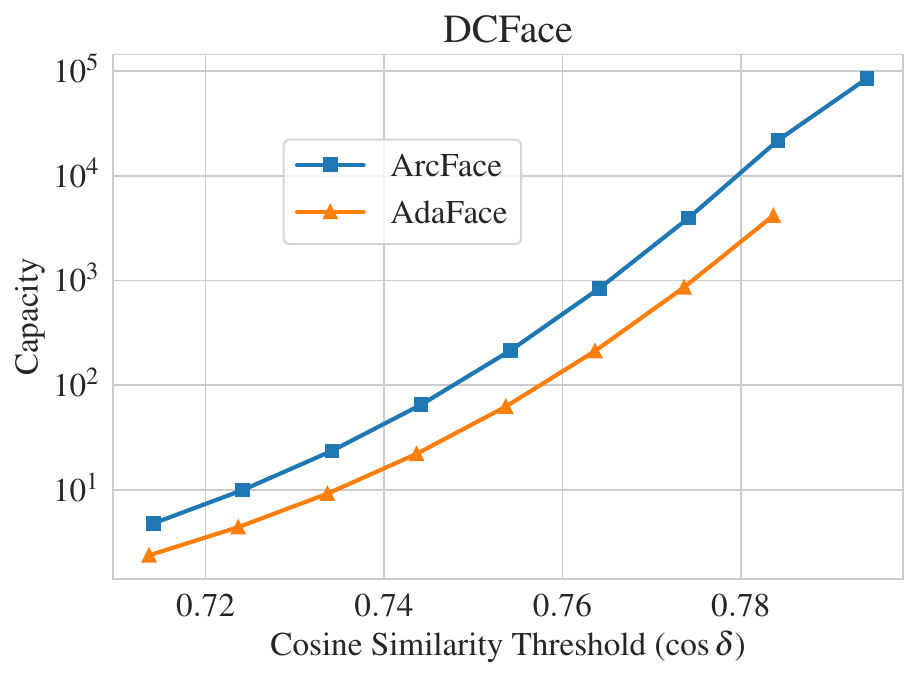}
    \end{subfigure}
    \caption{Effect on the choice of feature extractor on the capacity estimates. The capacity is slightly lower for AdaFace compared to ArcFace, but the trend is identical. Exact capacity values can be found at the \href{https://github.com/human-analysis/capacity-generative-face-models}{project page}.\label{fig:ablation-feature-extractors}}
    \vspace{-0.25cm}
\end{figure*}

\section{Ablation Study}
Here, we study the effect of our design choices on the stability and reliability of the capacity estimates.
\subsection{Effect of reference dataset choice\label{subsec:ref-dataset-ablation}}
As alluded to in Section~\ref{sec:intra-class-variance}, our capacity estimation algorithm for unconditional generators relies on the intra-class cosine distance of the face feature extractor. We used LFW as the reference dataset to determine the intra-class variance for the results presented in the previous section. Here we consider four other commonly used face verification benchmark datasets: (1)~\textbf{CFP-FP}~\cite{sengupta2016frontal}, (2)~\textbf{CALFW}~\cite{zheng2017crossage}, (3)~\textbf{AgeDB}~\cite{moschoglou2017agedb}, (4)~\textbf{CPLFW}~\cite{zheng2018crosspose} and utilize the threshold corresponding to 0.1\% FAR as a proxy for intra-class variance.

Figure~\ref{fig:ref-dataset-face-model-ablation} shows the trend in the capacity estimate due to variation of the reference dataset for each feature extractor. Observe that the capacity estimate at FAR of 0.1\% varies with the reference dataset's choice to determine the intra-class variance. We hypothesize that this phenomenon is due to the varying levels of intra-class diversity across the different datasets. As expected, the capacity of DCFace does not depend on the reference dataset since it is a conditional generator, and its intra-class variance is directly estimated from the class-conditional samples.

\begin{figure}[!ht]
    \vspace{-0.25cm}
    \centering
    \begin{subfigure}{0.23\textwidth}
        \centering
        \includegraphics[width=\textwidth]{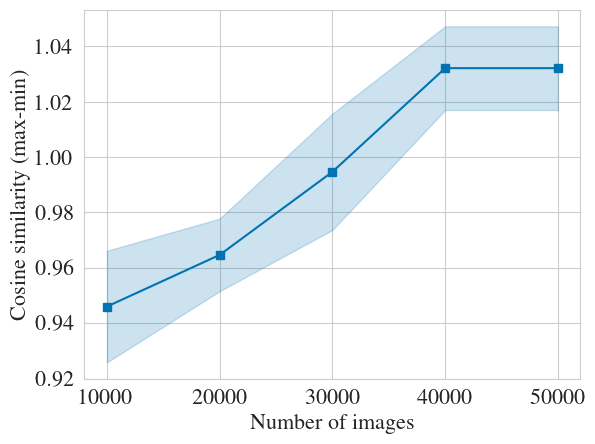}
        \caption{Cosine Similarity Range\label{fig:num-images-range}}
    \end{subfigure}
    \begin{subfigure}{0.23\textwidth}
        \centering
        \includegraphics[width=\textwidth]{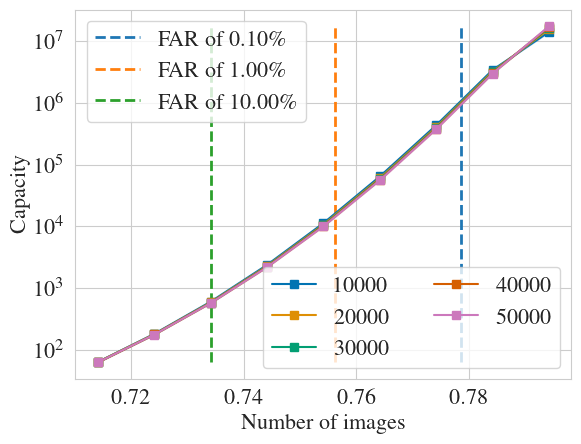}
        \caption{Capacity\label{fig:num-images-capacity}}
    \end{subfigure}
    \caption{Variation with the number of generated images\label{fig:num-images}}
    \vspace{-0.25cm}
\end{figure}
\vspace{3pt}
\subsection{Effect of sample size\label{subsec:num-images}}
Another design choice is the number of images from the generative models necessary for estimating the capacity. Ideally, we should be able to estimate it without exhaustively generating many images. Therefore, we estimate capacities from smaller subsets of the 50,000 images we generated for StyleGAN2+ADA and compare them against each other. Figures~\ref{fig:num-images-range} and~\ref{fig:num-images-capacity} show population variance and capacity estimates, respectively, as a function of the number of images. We observe a slight variation in the population variance as more images are used. However, the capacity estimates are stable even with only 10,000 images. In our approach, the capacity estimate depends on the population and intra-class variance, of which, for unconditional generators, only the former is affected by the number of generated images used to estimate the population variance.

\vspace{3pt}
\subsection{Effect of feature extractor choice\label{subsec:num-images}}
Finally, we looked at how the choice of feature space affected the representation of generated images and their estimated capacity. We tested two advanced models, ArcFace~\cite{deng2019arcface} and AdaFace~\cite{kim2022adaface}, to accomplish this. Figure~\ref{fig:ablation-feature-extractors} shows the capacity of the two feature extractors for various generative models. We observe that AdaFace has a slightly lower capacity than ArcFace, but both models share the same trend across all the generative models. This is expected since the difference between ArcFace and AdaFace is primarily the margin of separation between image identities based on the quality of the images. For high-quality images, both models perform similarly in terms of verification performance as well as capacity.

%% file: 06-conclusion.tex
\section{Concluding Remarks}

There are three aspects of interest in generative face models: \emph{photorealism} (resolution and facial details), \emph{diversity} (geometric, photometric, and demographic variations), and \emph{uniqueness} (number of distinct identities). While significant attention has been paid to photorealism and, to an extent, diversity, no attention has been paid to the uniqueness problem addressed in this paper. The generated face images were represented in a hyperspherical space of a feature extractor, and an exact formula for estimating an upper bound on the capacity as a function of a desired false acceptance rate in this space was presented. Empirically, we estimated the capacity of multiple generative face models across demographic attributes like age and gender. For the StyleGAN family of models, numerical results yielded a capacity of 1.43 million at a FAR of 0.1\%, which drops quickly to 562 at 10\%. The generative face models did not exhibit any disparities in capacity w.r.t gender, while some models exhibited capacity variations across different age groups. Finally, we demonstrated that our capacity estimates are robust to the number of image samples and choice of feature space.

As generative face models make rapid strides in their photorealism and witness wider adoption, quantifying their capacity is a significant problem, both from an analytical and practical perspective. However, due to the challenging nature of finding a closed-form expression for capacity, we represent the images in the hyperspherical space of a state-of-the-art feature extractor and make simplifying assumptions on the distribution of the intra-class variance in that space. Experimental results demonstrate that our approach can provide reasonable capacity estimates. Relaxing the assumptions of the approach presented here is an exciting direction for future work, leading to even tighter capacity estimates.

\vspace{3pt}
\noindent\textbf{Acknowledgments:} This material is based upon work supported by the Center for Identification Technology Research and the National Science Foundation under Grant No. 1841517.